\documentclass[10pt,twocolumn,letterpaper]{article}

\usepackage{cvpr}              % To produce the CAMERA-READY version

\usepackage{colortbl}
\usepackage{xcolor}

\usepackage[accsupp]{axessibility}

\definecolor{cvprblue}{rgb}{0.21,0.49,0.74}
\usepackage[pagebackref,breaklinks,colorlinks,citecolor=cvprblue]{hyperref}

\def\confName{CVPR}
\def\confYear{2024}

\title{ImageNet-D: Benchmarking Neural Network Robustness on \\ Diffusion Synthetic Object}

\newcommand{\samethanks}[1][\value{footnote}]{\footnotemark[#1]}
\author{Chenshuang Zhang$^{1}$~~~~
Fei Pan$^{2}$~~~~Junmo Kim$^{1}$\thanks{Corresponding author. Junmo Kim $<$junmo.kim@kaist.ac.kr$>$,  Chengzhi Mao $<$chengzhi.mao@mila.quebec$>$.}~~~~ In So Kweon$^{1}$~~~~~~
Chengzhi Mao$^{3,4}$\samethanks\\
KAIST$^{1}$, University of Michigan, Ann Arbor$^{2}$,   McGill University$^{3}$, MILA$^{4}$ \\
}

\begin{document}
\maketitle

\begin{abstract}
We establish rigorous benchmarks for visual perception robustness. Synthetic images such as ImageNet-C, ImageNet-9, and Stylized ImageNet provide specific type of evaluation over synthetic corruptions, backgrounds, and textures, yet those robustness benchmarks are restricted in specified variations and have low synthetic quality. In this work, we introduce generative model as a data source for synthesizing hard images that benchmark deep models' robustness. Leveraging diffusion models, we are able to generate images with more diversified backgrounds, textures, and materials than any prior work, where we term this benchmark as ImageNet-D.  Experimental results show that ImageNet-D results in a significant accuracy drop to a range of vision models, from the standard ResNet visual classifier to the latest foundation models like CLIP and MiniGPT-4, significantly reducing their accuracy by up to 60\%. Our work suggests that diffusion models can be an effective source to test vision models. The code and dataset are available at \small{\url{https://github.com/chenshuang-zhang/imagenet_d}}.

\end{abstract}
\section{Introduction}
\label{sec:intro}

Neural networks have achieved remarkable performance in tasks ranging from image classification~\cite{vaswani2017attention,liu2021swin,liu2022convnet} to visual question answering~\cite{li2023blip,dai2023instructblip,liu2023visual,zhu2023minigpt}. These advances have inspired the application of neural networks in  various fields, including security and safety-critical systems such as self-driving cars~\cite{kangsepp2022calibrated,nesti2023ultra,liu2023vectormapnet}, malware detection~\cite{yuan2014droid,chen2019believe,pei2017deepxplore} and  robots ~\cite{brohan2022rt,brohan2023rt,huang2023voxposer}.  Due to their wide adaptation, it is becoming increasingly important to identify the robustness of neural networks~\citep{ming2022delving,li2023distilling} for safety reasons.

To evaluate the robustness of neural networks, ObjectNet~\cite{barbu2019objectnet} collects real-world object images on controlled factors like background with human workers, which is time-consuming and labor-intensive. To scale up data collection, synthetic images are proposed as test images~\cite{geirhos2018imagenet,hendrycks2019benchmarking,xiao2020noise}. For example, ImageNet-C~\cite{hendrycks2019benchmarking} introduces a set of low-level common visual corruptions, such as gaussian noise and blur, to test models' robustness. ImageNet-9~\cite{xiao2020noise} uses simple cutting and paste technique to create robustness benchmark on object background, yet the images are not realistic. Stylized-ImageNet~\cite{geirhos2018imagenet} generates new images by altering the textures of ImageNet images, which cannot control the global factors like background.

\begin{figure}[!tbp]
    \centering
        \includegraphics[width=1.0\linewidth]{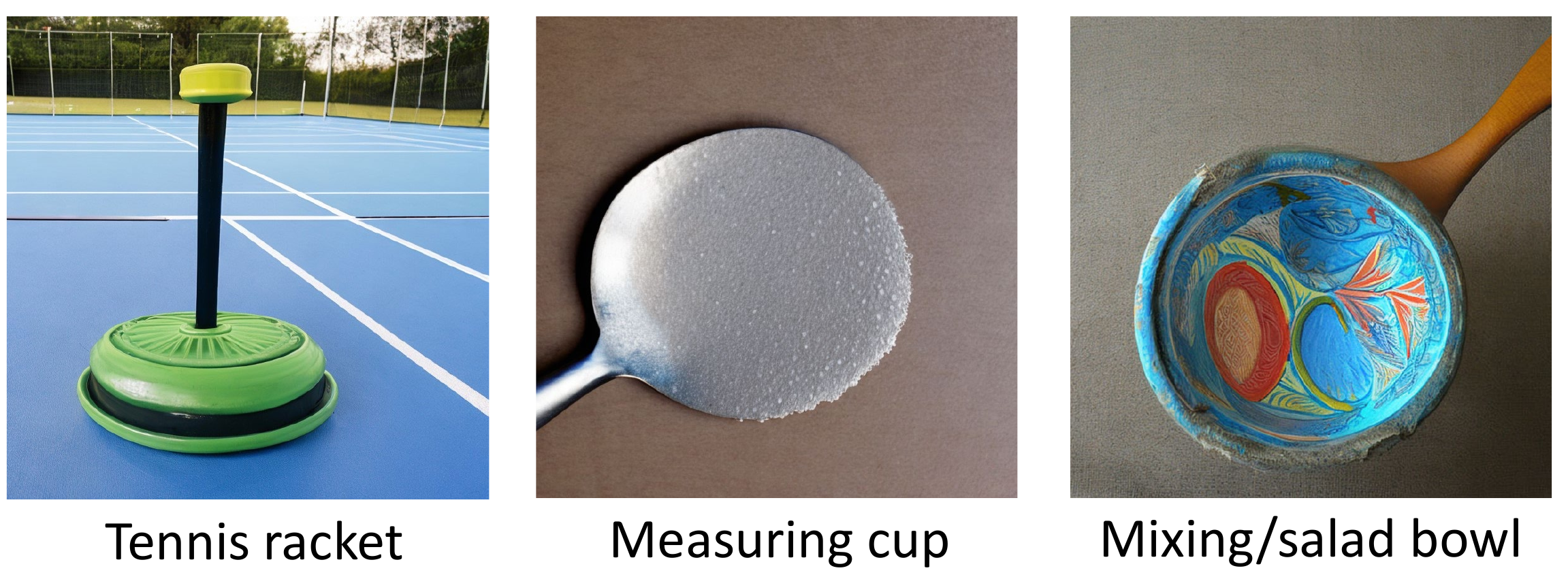}
    \caption{Top predictions from CLIP (ViT-L/14) on ImageNet-D. We synthesize the images by changing their background, texture and material. The groundtruth for the images are \textit{plunger}, \textit{spatula}, and \textit{ladle} in order, together with the background (badminton court), texture (freckled), and material (painted). 
      }
      \vspace{-5mm}
    \label{fig:teaser_figure}
\end{figure}

In this work, we introduce ImageNet-D, a synthetic test set generated by diffusion models for object recognition task. Capitalizing on the capability of pioneering Stable Diffusion models\cite{rombach2022high}, we show that we can steer diffusion models with language to create realistic test images that cause vision models fail. Figure~\ref{fig:teaser_figure} shows three
failure cases of CLIP model on our synthetic ImageNet-D dataset.
Since we rely on language to create images, we can vary the high-level factors in the images in contrast to the local corruptions and texture in prior work, introducing addition factors that one can evaluate robustness on.

To enhance  sample difficulty of our dataset, we selectively retain images that cause failures in multiple chosen vision models. Our results show that images triggering errors in chosen  models can reliably transfer their challenging nature to other, previously untested models. This leads to a notable decrease in accuracy, even in state-of-the-art foundation models like MiniGPT-4~\cite{zhu2023minigpt} and LLaVa~\cite{liu2023visual}, suggesting our dataset reveals common failures in vision models.

\begin{figure}[!tbp]
    \centering
        \includegraphics[width=1.0\linewidth]{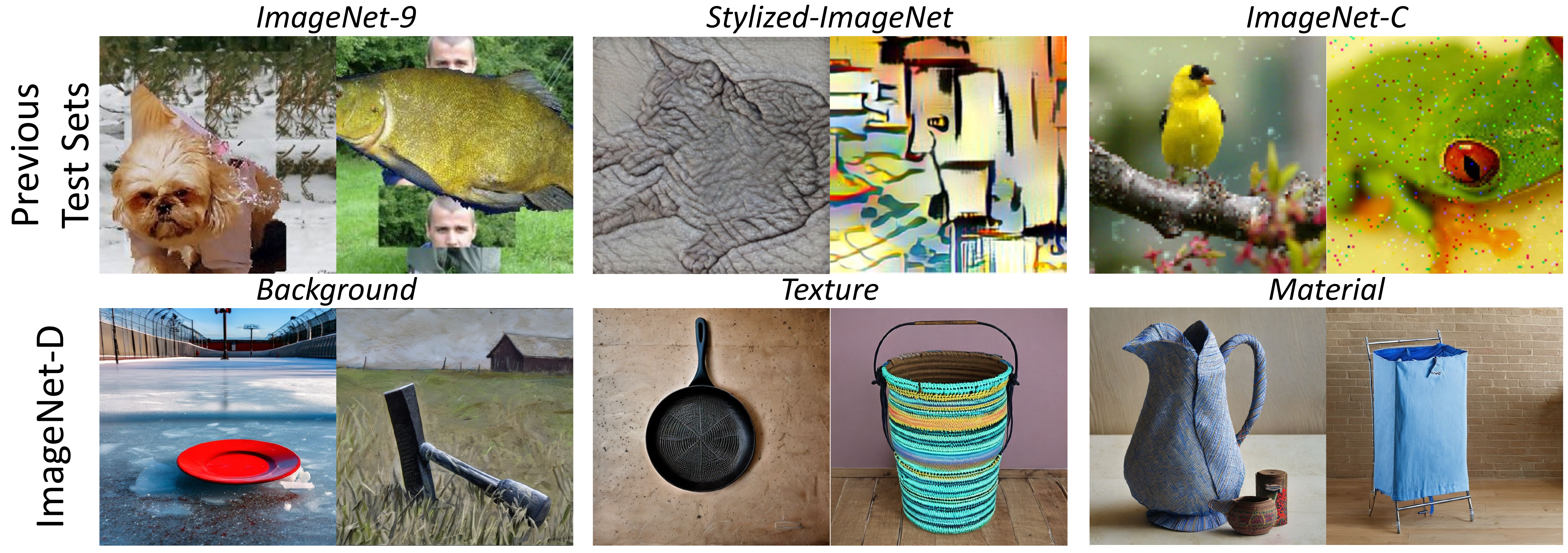}
    \caption{Examples from  ImageNet-9~\cite{xiao2020noise}, Stylized-ImageNet~\cite{geirhos2018imagenet} and ImageNet-C~\cite{hendrycks2019benchmarking} and our ImageNet-D. For the second row, we show images from ImageNet-D with different backgrounds, textures and materials orderly. Take the background for example (the two columns on the left), ImageNet-9~\cite{xiao2020noise} generates new images by simply cutting and paste foreground and background from different images, leading to object deformation and dislocation. By contrast, ImageNet-D includes images with diverse backgrounds by diffusion generation, achieving superior visual fidelity. 
    }
    \label{fig:test_set_comparison}
\end{figure}

Visualizations demonstrate that Imagenet-D significantly enhances image quality compared to previous synthetic robustness benchmarks, as evidenced in Figure~\ref{fig:test_set_comparison}. Imagenet-D serves as an effective tool for reducing the performance and assessing model robustness,  including ResNet 101 (reducing 55.02\%), ViT-L/16 (reducing 59.40\%),  CLIP (reducing 46.05\%), and transfer well to unforeseen large vision language models like LLaVa~\cite{liu2023visual} (reducing 29.67\%), and MiniGPT-4~\cite{zhu2023minigpt} (reducing 16.81\%). Our approach of utilizing generative models to evaluate model robustness is general, and shows significant potential for even greater effectiveness with future advancements in generative models.

\begin{figure*}[!tbp]
    \centering
        \includegraphics[width=1.0\linewidth]{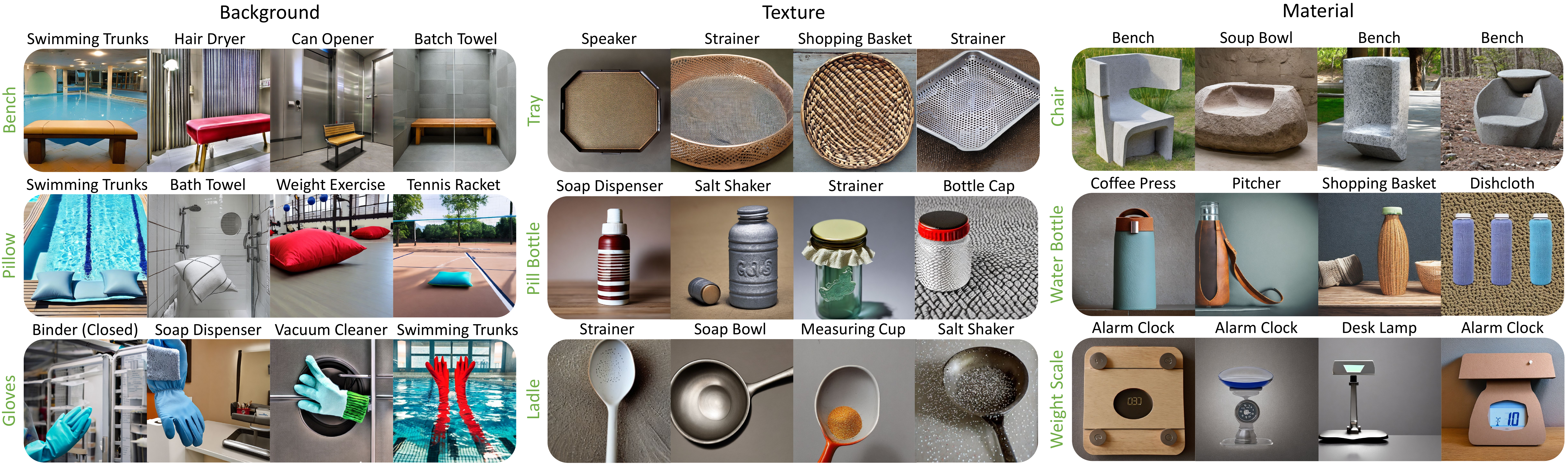}
    \caption{The ImageNet-D test set. Each group of images is generated with the same object and nuisance, such as background, texture, and material. For each group of images, the ground truth label is color green, while the predicted categories by CLIP (ViT-L/14) on each image are in black. Leveraging diffusion models for image generation, we can create a test set with diverse combinations of objects and nuisances. For example, the top left corner shows a bench in the swimming pool background. Interestingly, CLIP (ViT-L/14) recognizes the bench in this image as swimming trunks.
    } 
    \label{fig:dataset_examples}
\end{figure*}   

\section{Related work}
\label{sec:related_work}

\noindent \textbf{Robustness of neural networks.} Neural networks have evolved from CNN~\cite{he2016deep,huang2017densely}, ViT~\cite{vaswani2017attention,liu2021swin},  to large foundation models~\cite{bommasani2021opportunities,devlin2018bert,touvron2023llama}.  Previous work has investigated neural networks robustness from multiple aspects, such as  adversarial examples~\cite{mao2022understanding, mahmood2021robustness,madry2017towards,zhao2023evaluating,zhang2019theoretically} and out-of-domain samples~\cite{MAE, mao2021discrete,hendrycks2021many, augmix}. Foundation models have shown greater robustness on out-of-distribution samples~\cite{radford2021learning}. Robust explanation has also been investigated~\cite{mao2023doubly, liu2023visual, zhu2023minigpt}. To systematically evaluate the robustness of deep models, test sets that cover different factors are urgently needed.

\noindent \textbf{Dataset for benchmarking robustness.}  To evaluate neural network robustness,  a branch of studies source images online, including  ImageNet-A~\cite{hendrycks2021natural}, Imagenet-R~\cite{hendrycks2021many} and ImageNet-Sketch~\cite{wang2019learning}. However, they are limited to images that exist on the web. ObjectNet~\cite{barbu2019objectnet} manually collects images  with the help of 5982 workers, which is time-consuming and resource-intensive.

To overcome the limitations of web images and reduce the cost of manual collection, synthetic images are proposed for robustness evaluation~\cite{geirhos2018imagenet,hendrycks2019benchmarking,xiao2020noise}. ImageNet-C ~\cite{hendrycks2019benchmarking} benchmarks model robustness on low-level corruptions. ImageNet-9~\cite{xiao2020noise} generates new images by combining foreground and background from different images, however, limited by poor image fidelity. Stylized-ImageNet~\cite{geirhos2018imagenet} alters the textures of ImageNet images by AdaIN style transfer~\cite{huang2017arbitrary} or introducing texture-shape cue conflict, which cannot control other factors like backgrounds. In this work, we introduce a new test set ImageNet-D, which is generated by controlling diffusion models and includes novel images with diverse backgrounds, textures, and materials.

\noindent \textbf{Image generation.} Diffusion models have achieved great success in various tasks including image generation~\citep{saharia2022photorealistic,ramesh2022hierarchical,ruiz2023dreambooth,zhang2023text}.
As a milestone work, Stable diffusion~\citep{rombach2022high} enables high-fidelity image synthesis controlled by language. InstructPix2Pix~\cite{brooks2023instructpix2pix} provides a more sophisticated control by editing a given image according to human instructions.  In this paper, we build our pipeline with the standard Stable Diffusion model, yet our algorithm is compatible with other generative models that can be steered by language.

\noindent \textbf{Enhancing perception with diffusion images}. Diffusion-generated images have been used for vision perception tasks. A branch of 
 studies~\cite{yuan2023not,bansal2023leaving,azizi2023synthetic,tian2023stablerep} 
 improves classification accuracy by using synthetic images as training data augmentation. DREAM-OOD~\cite{du2023dream} finds the outliers by decoding sampled latent embeddings to images. However, their method lacks specific control over image space, which is crucial for benchmarks like ImageNet-D. ~\cite{metzen2023identification} identifies under-represented attribute pairs, while our study focuses on hard images with a single attribute.  Unlike ~\cite{li2023imagenet,vendrow2023dataset,prabhu2023lance} that modify existing datasets, our work generates new images and mines the most challenging ones as the test set, achieving greater accuracy drop than~\cite{li2023imagenet,vendrow2023dataset,prabhu2023lance}.

\section{ImageNet-D}
\label{sec:imagenet_d}

We first present how ImageNet-D is created in Section~\ref{sec:dataset_design}, followed by an overview of its statistics in Section~\ref{sec:statistics}.

\begin{figure*}[!tbp]
    \centering
    \includegraphics[width=1.0\linewidth]{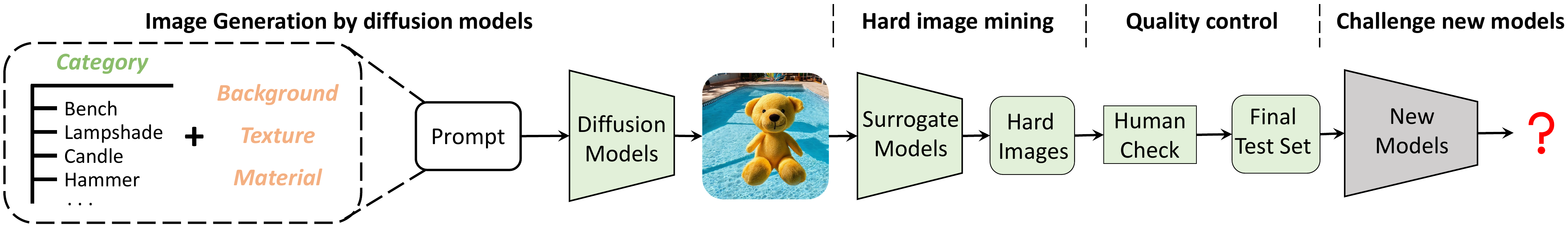}
    \caption{ImageNet-D creation framework. ImageNet-D is created by first combining various object categories and nuisances, including background, texture, and material. 
    To make the test set challenging, we only keep the hard images from the large pool that commonly make multiple surrogate models fail to predict the correct object label. The test set is then refined through human verification to ensure the images are valid, single-class, and high-quality, making ImageNet-D suitable for evaluating the robustness of different neural networks.
    }
    \label{fig:framework}
\end{figure*}

\subsection{Dataset design}
\label{sec:dataset_design}

While neural networks excel in various applications, their robustness needs rigorous evaluation for safety. Traditional evaluations use existing test sets, including either natural  ~\cite{barbu2019objectnet,hendrycks2021natural} or synthetic images~\cite{geirhos2018imagenet,hendrycks2019benchmarking,xiao2020noise}. Compared to manual image collection, collecting a synthetic test set is more efficient~\cite{geirhos2018imagenet,xiao2020noise}. However, the diversity of current synthetic test sets is limited due to their reliance on existing images for attribute extraction. These synthetic images are also not realistic, as shown in Figure~\ref{fig:test_set_comparison}. ImageNet-D is introduced to evaluate model robustness across various object and nuisance combinations, addressing these limitations.

\textbf{Image generation by diffusion models.} To construct ImageNet-D, diffusion models are used to create a vast image pool by combining all possible object and nuisances, enabling high-fidelity image generation based on user-defined text inputs.
We use Stable Diffusion model\cite{rombach2022high} for image generation, while our approach is compatible with other generative models that can be steered by language. The image generation process is formulated as follows:
\begin{equation}
   \text{Image}(C, N)  = \text{Stable Diffusion}(\text{Prompt}(C,N)),
    \label{eq:image_generation}
\end{equation}
where $C$ and $N$ refer to the object category and nuisance, respectively. The nuisance $N$ includes background, material, and texture in this work. Table~\ref{tab:prompt_list} presents an overview of nuisances and prompts to diffusion models. Using the backpack category as an example, we first generate images of backpacks with various  backgrounds, materials, and textures (e.g., a backpack in a wheat field), offering a broader range of combinations than existing test sets. Each image is labeled with its prompt category  $C$  as  ground truth for classification. An image is viewed misclassified if the model's predicted label does not match the ground truth $C$.

\begin{table*}[!htbp]
    \centering    
    \caption{Overview of nuisances and input prompts to diffusion models.
    During ImageNet-D construction, 468 backgrounds, 47 textures, and 32 materials from the Broden dataset~\cite{bau2017network} are used as nuisances. Images are generated by pairing each object with all nuisances in diffusion model prompts. This approach allows for efficient scaling of ImageNet-D with additional categories and nuisances.
    }
    \label{tab:prompt_list}
    \resizebox{1.0\linewidth}{!}{ 
    \begin{tabular}{l|c|l|lccccccccccc}
    \toprule 
       Nuisance & Nuisance number & Prompt to diffusion models & Prompt example\\
     \toprule
    Background & 468  &  A $\left[ \texttt{category}\right]$ in the $\left[ \texttt{background} \right]$ & A \texttt{backpack} in the \texttt{wheat field}\\
    Texture &   47  & A $\left[ \texttt{texture}\right]$ $\left[ \texttt{category}\right]$ & A \texttt{knitted}  \texttt{backpack}\\
   Material   & 32&  A $\left[ \texttt{category}\right]$ made of $\left[ \texttt{material}\right]$ & A \texttt{backpack} made of \texttt{leather}\\  

    \bottomrule 
    \end{tabular}}
\end{table*}

\begin{table}[!tbp]
    % \vspace{-14pt}
    \centering
    \caption{Test accuracy of CLIP (ViT-L/14) on the synthetic image pool  by \textbf{exhausting all} the 
    object category and nuisance combinations.  We show that CLIP achieves high accuracy on the synthetic image pool. To create a challenging test set for robustness evaluation,  we further mine the hard samples as the final test set.}
    \label{tab:vanilla_generation}
    \resizebox{1.0\linewidth}{!}{ 
    \begin{tabular}{ccccccc}
    \toprule 
  Test Set   &ImageNet&ObjectNet & \multicolumn{3}{c}{ \cellcolor[gray]{0.9} Synthetic image pool} \\
     &   &  &  \cellcolor[gray]{0.9}Background &  \cellcolor[gray]{0.9}Texture & \cellcolor[gray]{0.9} Material  \\   
    \toprule
   Acc (\%) &  74.64 & 66.91	 & \cellcolor[gray]{0.9} 95.79 &  \cellcolor[gray]{0.9} 94.02 & \cellcolor[gray]{0.9} 93.75  \\
    \bottomrule 
    \end{tabular}}
\end{table}

After creating a large image  pool with all object category and nuisance pairs, we evaluate CLIP (ViT-L/14) model on  these images in Table~\ref{tab:vanilla_generation}. Experimental details  are reported in  Section~\ref{sec:experimental_setup}. Table~\ref{tab:vanilla_generation} shows that CLIP achieves high accuracy on all the test sets, with an accuracy of around 94\% on synthetic image pool. To create a challenging test set for robustness evaluation, we propose an efficient strategy to find the hard test samples from all generated images as follows.

\begin{figure*}[!htbp]
    \centering
    \includegraphics[width=1.0\linewidth]{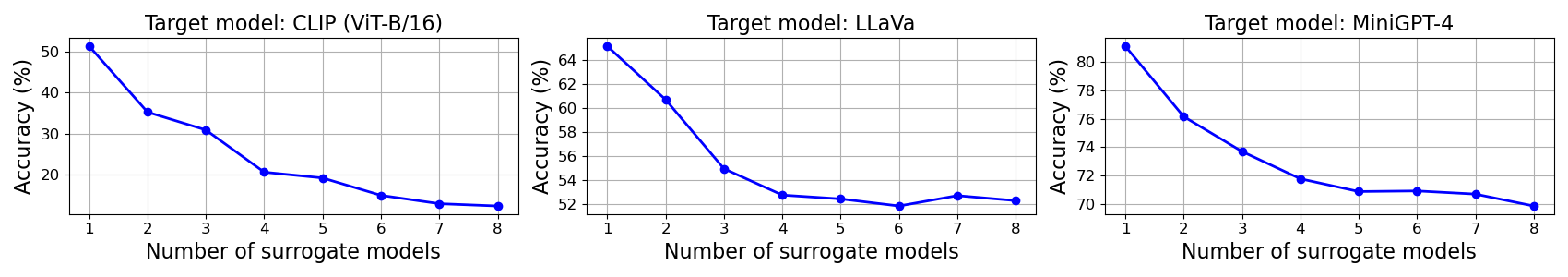}
    \caption{Test accuracy of target models on shared failures of surrogate models. We adopt  known surrogate models to identify their shared failure images as the test set, then evaluate a new target model on this test set.  We show that the shared failures of surrogate models can also deceive a new target model, leading to low test accuracy. Increasing the number of surrogate models lowers the target models' test accuracy, suggesting a more difficult test set. 
    }
    \label{fig:filter_consistency}
\end{figure*}

\textbf{Hard image mining with shared perception failures.} Before introducing how to identify hard samples from synthetic image pool,  we first define the concept of shared perception failure as follows.

\texttt{Shared failure:} An image is a shared failure if it leads multiple models to predict  object's label incorrectly.

An ideal hard test set should include images that all tested models fail to identify. However,   this is impractical due to the inaccessibility of future test models, termed target model. Instead, we construct the test set using shared failures of known surrogate models. If the failures of surrogate models lead to low accuracy in unknown models, the test set is deemed challenging. This is defined as transferable failure as follows:

\texttt{Transferable failure:} Shared failures of known surrogate models are transferable if they also result in low accuracy for unknown target models.

To test if shared failures of surrogate models are transferable for diffusion-generated images, we evaluate test sets created with shared failures from 1 to 8 surrogate models  in  Figure~\ref{fig:filter_consistency}.  We evaluate the accuracy of three target models that are not used during test set construction, including CLIP(ViT-B/16), LLaVa, and MiniGPT-4.  Figure~\ref{fig:filter_consistency} shows that target model accuracy decreases as more surrogate models are used. The test sets are created with diverse backgrounds, while experiments for texture and material show the same trend.   This trend demonstrates that failure images of multiple surrogate models can form a challenging test set for unseen new models. Notably, the accuracy decrease slows when the number of surrogate models exceeds four.

\textbf{Quality control by human-in-the-loop.}  The above process allows us to automatically find a challenging test set to unseen models. However, generative models can produce incorrect images not matching the prompt category.  We resort to human annotation to 
ensure the ImageNet-D images are simultaneously valid, single-class, and high-quality. After first-round annotation by graduate students, we use Amazon Mechanical Turk~\cite{deng2009imagenet,recht2019imagenet,hendrycks2021many} to evaluate labeling quality. We ask the workers to select the images that they can either recognize the main object or the main object can be used functionally as the ground truth category. Moreover, we design sentinels to ensure high-quality responses, including positive, negative and consistent sentinels. We report details of the labeling task in the appendix. A total of 679 qualified workers participated in 1540 labeling tasks, achieving an agreement of 91.09\%. Figure~\ref{fig:dataset_examples} displays images from ImageNet-D, demonstrating high fidelity and diversity in object and nuisance pairs. We summarize the framework of creating ImageNet-D in Figure~\ref{fig:framework}.

\subsection{Dataset statistics}
\label{sec:statistics}

\begin{figure*}[!htbp]
    \centering
    \begin{minipage}{.30\textwidth}
        \centering
       \includegraphics[width=1.0\textwidth]{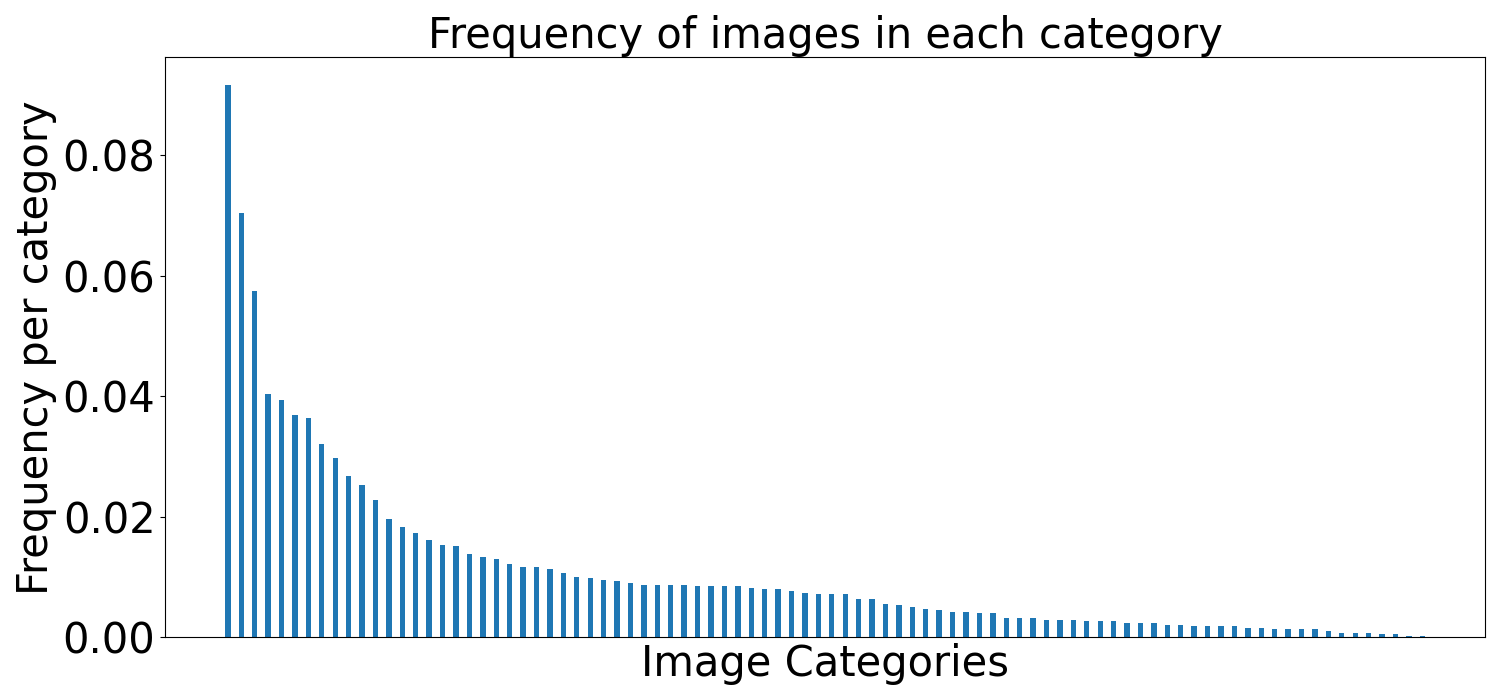}
        \subcaption{Background}
    \end{minipage}
    \begin{minipage}{.30\textwidth}
        \centering
     \includegraphics[width=1.0\textwidth]{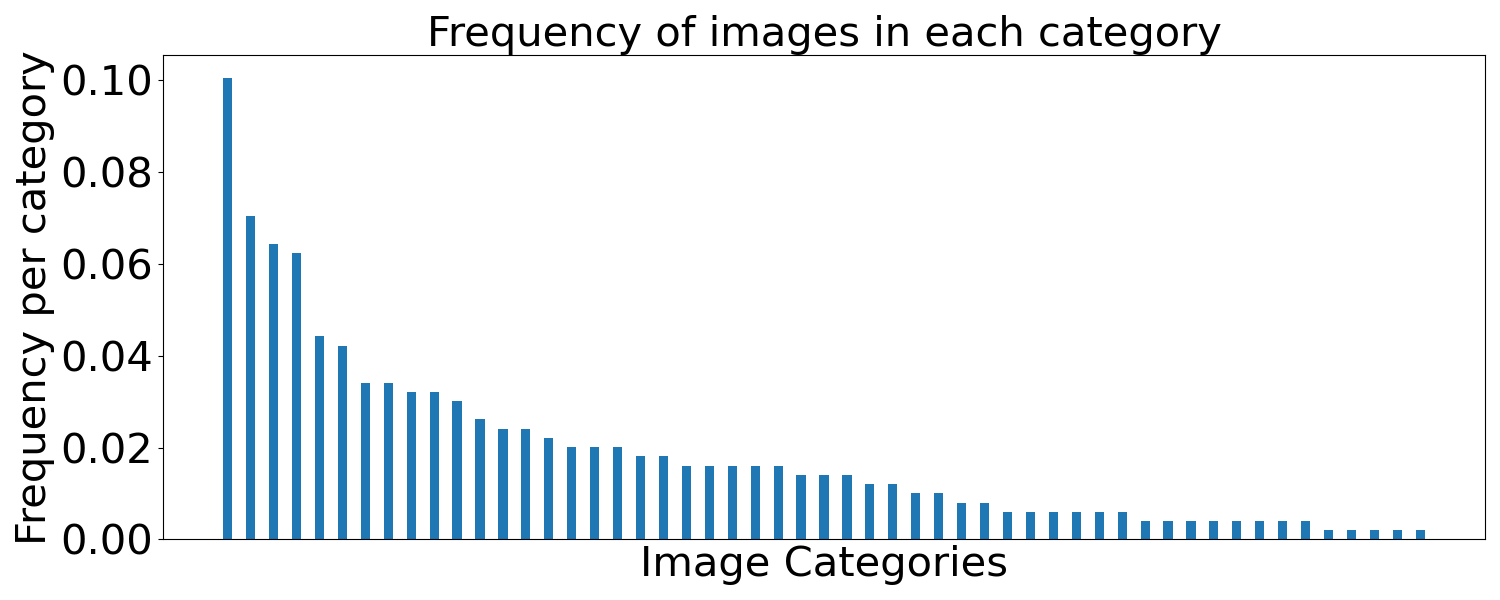}
      \subcaption{Texture}
    \end{minipage}
    \begin{minipage}{.30\textwidth}
        \centering
     \includegraphics[width=1.0\textwidth]{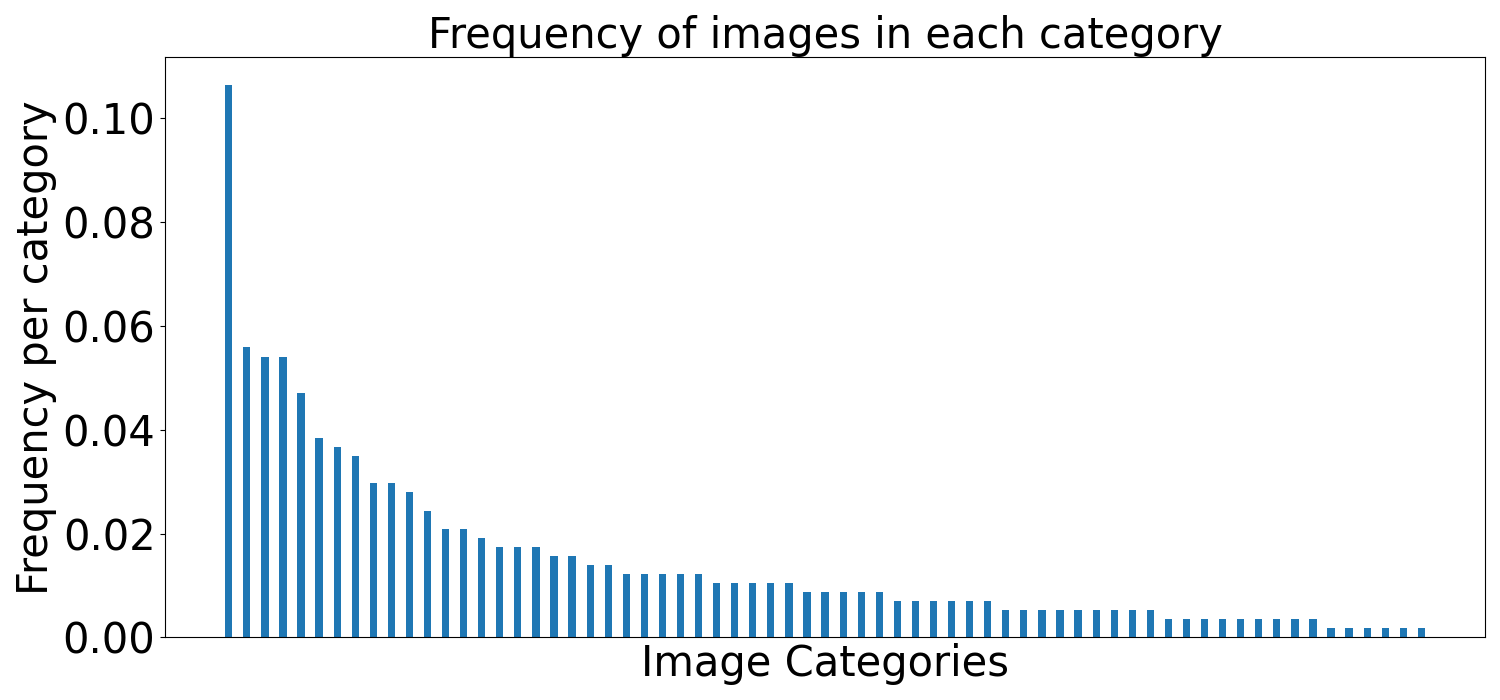}
         \subcaption{Material}
    \end{minipage}
    \caption{Histogram of the image frequency per category in our test set, following a natural long-tail distribution.
    }
    \label{fig:hist_category}
\end{figure*}

\begin{figure*}[!htbp]
    \centering
        \begin{minipage}{.60\textwidth}
        \centering
       \includegraphics[width=1.0\textwidth]{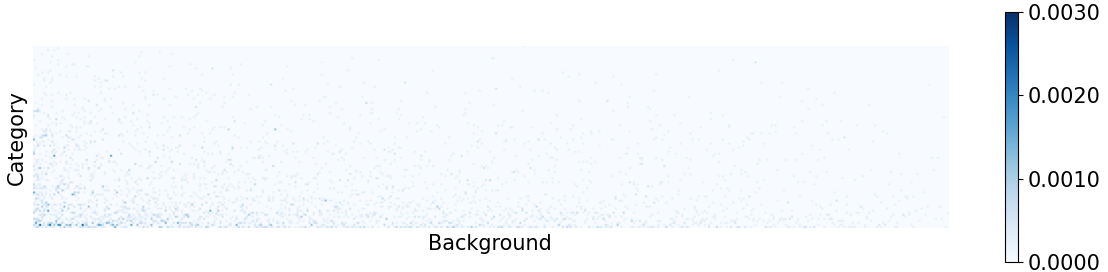}
        \subcaption{Background}
    \end{minipage}
        \begin{minipage}{.16\textwidth}
        \centering
     \includegraphics[width=1.0\textwidth]{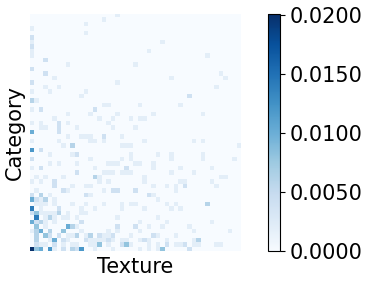}
      \subcaption{Texture}
    \end{minipage}
    \begin{minipage}{.11\textwidth}
        \centering
     \includegraphics[width=1.0\textwidth]{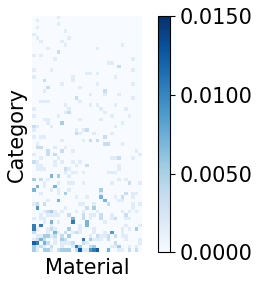}
         \subcaption{Material}
    \end{minipage}

    \caption{Frequency of object category and nuisance pairs. Each dot indicates a pair, while the x-axis and y-axis indicate the nuisance and category, respectively. As shown in Table~\ref{tab:prompt_list},  we adopt 468 backgrounds, 47 textures, and 32 materials from the Broden dataset~\cite{bau2017network}, leading to different widths of Figure (a) (b) (c). A darker color indicates more samples. The sparse and non-uniform distribution shows the necessity of exhausting all category and nuisance pairs in ImageNet-D creation.
    }
    \label{fig:heatmap}
\end{figure*}

ImageNet-D includes 113 overlapping categories between ImageNet and ObjectNet, and 547 nuisances candidates from the Broden dataset~\cite{bau2017network}(see Table~\ref{tab:prompt_list}), resulting in 4835 hard images featuring diverse backgrounds (3,764), textures (498), and materials (573).  Our pipeline to create ImageNet-D  is general  and efficient, allowing easy addition of new categories and nuisances. ImageNet-D's category  distribution exhibits a natural long-tail pattern, as shown in Figure~\ref{fig:hist_category}. The sparse and  non-uniform category-attribute distribution in Figure~\ref{fig:heatmap} shows the necessity of exhausting all category and nuisance pairs in test set creation.

\section{Experiments}
\label{sec:benchmark}

We evaluate various models on ImageNet-D benchmark. We find that ImageNet-D significantly decreases all models' accuracy by up to  60\%.  We then show whether prior advancements could improve ImageNet-D robustness, such as data augmentation. Lastly, we discuss ImageNet-D from various aspects, such as nearest neighbor retrieval.

\subsection{Experimental setups}
\label{sec:experimental_setup}

\textbf{Test set construction setups.} We use Stable Diffusion~\cite{rombach2022high} to create ImageNet-D, and  adopt the pretrained weight of version stable-diffusion-2-1 from Hugging Face. To find the hard images, we finalize ImageNet-D with shared failures of 4 surrogate models, including CLIP~\cite{radford2021learning} (ViT-L/14, ViT-L/14-336px and ResNet50), and vision model (ResNet50~\cite{he2016deep}). The candidate set of surrogate models in Figure~\ref{fig:filter_consistency} also includes CLIP (Resnet101,ViT-B/32) and vision model (ViT-L/16~\cite{dosovitskiy2010image} and VGG16~\cite{simonyan2014very}).

\textbf{Evaluation of classification models.} Robustness on ImageNet-D is measured by top-1 accuracy in object recognition, the ratio of correctly classified images to total images. We evaluate classification models with the open-source pretrained weights.  For CLIP~\cite{radford2021learning}, we follow the original paper~\cite{radford2021learning} to adopt  
\textit{A  photo of a $\left[ category \right]$} as the text template.  The zero-shot accuracy of CLIP is reported.

\textbf{Evaluation of visual question answering (VQA) models.} We evaluate the accuracy of the state-of-the-art 
open-source VQA models on ImageNet-D, including LLaVa~\cite{liu2023visual}, and MiniGPT-4~\cite{zhu2023minigpt}. Given an input image, VQA models  output answers based on input text prompt.  However, the textual output of VQA models is not limited to a certain template,  thus may not include the category name in pre-defined category list of object recognition tasks. This makes it hard to assess the accuracy based on diverse answers.

A common  prompt that asks  VQA models to recognize the object is: \texttt{What is the main object in this image?}  To make VQA models choose from  pre-defined category list, we ask VQA models as follows:  \texttt{What is the main object in this image? Choose from the following list: $\left[ \text{GT category} \right]$, $\left[ \text{failure category} \right]$}. GT category refers to the image's ground truth category. As for the failure category, we adopt the category that achieves the highest CLIP (ViT-L/14) confidence among all wrong categories. With this  prompt, we find that both MiniGPT-4 and LLaVa can choose from  provided category list in their output. 
If the model chooses ground truth category, this image is viewed to be correctly recognized. 
Therefore, we can compute the accuracy of VQA models.

\subsection{Robustness evaluation}
\label{sec:experimental_results}

\begin{figure*}[!htbp]
    \centering
    % \vspace{-5mm}
    \includegraphics[width=1.0\linewidth]{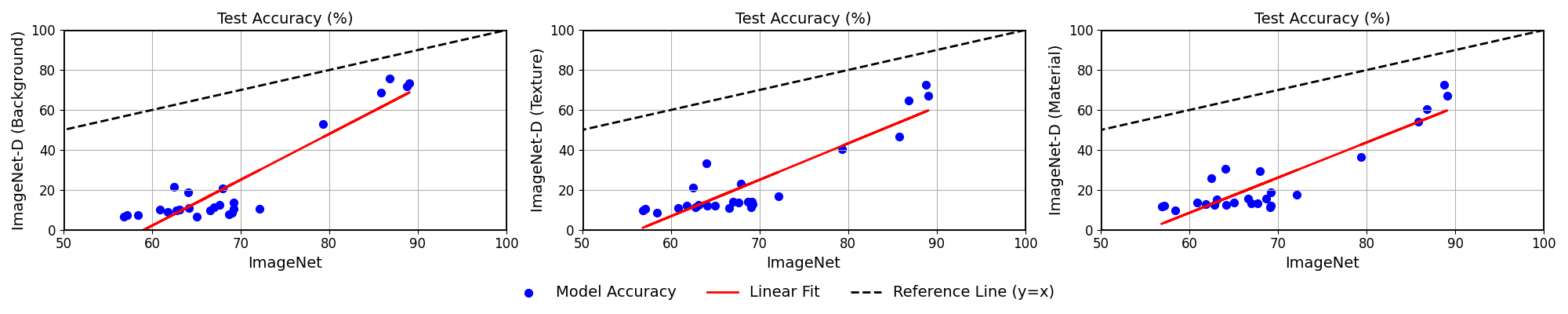}
    \caption{Model accuracy on ImageNet vs.  ImageNet-D. Each data point corresponds to one tested model. The plots reveal  that there is a significant accuracy drop from ImageNet to our new test set ImageNet-D. As the model's accuracy on ImageNet increases, the accuracy on ImageNet-D is also higher. These results show the effectiveness of ImageNet-D to evaluate the robustness of neural networks. We report the exact number of 14 models of this figure in Table~\ref{tab:benchmark_results},  and the results for all models can be found  in the appendix. 
    }
    \label{fig:main_result_figure}
\end{figure*}

\begin{table*}[!htbp]
    \centering
    \scriptsize
    \caption{Test accuracy of vision models and large foundation models (\%). We show the test accuracy for the vision models and large foundation models (rows) on different test sets (columns). The numbers in green refer to the accuracy drop of ImageNet-D compared to ImageNet. For MiniGPT-4 and LLaVa, ImageNet-D reduces the accuracy by 16.81\% and 29.67\% compared to the ImageNet, respectively. Our results show that ImageNet-D is effective to evaluate the robustness of neural networks.
    } 
    \label{tab:benchmark_results}
    \resizebox{1.0\linewidth}{!}{ 
    \begin{tabular}{ll|c|c|c|c|ccc|cccc}
    \toprule  
Model & Architecture &  ImageNet & ObjectNet & ImageNet-9 & Stylized &  \multicolumn{3}{c|}{\cellcolor[gray]{0.9}  ImageNet-D} & \cellcolor[gray]{0.9}   ImageNet-D \\
    && &  &   &  & \cellcolor[gray]{0.9} Background & \cellcolor[gray]{0.9} Texture & \cellcolor[gray]{0.9} Material & \cellcolor[gray]{0.9} Total\\
  \toprule 
      
 Vision model (CNN) &VGG19&62.77&27.19&74.84&16.25&\cellcolor[gray]{0.9}9.8({\color[HTML]{009901}-52.97})&\cellcolor[gray]{0.9}11.45({\color[HTML]{009901}-51.32})&\cellcolor[gray]{0.9}12.39({\color[HTML]{009901}-50.38})&\cellcolor[gray]{0.9}10.28({\color[HTML]{009901}-52.49})\\
&ResNet101&67.66&32.34&81.85&22.66&\cellcolor[gray]{0.9}12.38({\color[HTML]{009901}-55.28})&\cellcolor[gray]{0.9}13.65({\color[HTML]{009901}-54.01})&\cellcolor[gray]{0.9}13.44({\color[HTML]{009901}-54.22})&\cellcolor[gray]{0.9}12.64({\color[HTML]{009901}-55.02})\\
&Densenet161&66.99&31.86&84.91&22.5&\cellcolor[gray]{0.9}11.34({\color[HTML]{009901}-55.65})&\cellcolor[gray]{0.9}14.06({\color[HTML]{009901}-52.93})&\cellcolor[gray]{0.9}13.26({\color[HTML]{009901}-53.73})&\cellcolor[gray]{0.9}11.85({\color[HTML]{009901}-55.14})\\
&Wideresnet101&69.2&34.37&82.17&21.48&\cellcolor[gray]{0.9}10.55({\color[HTML]{009901}-58.65})&\cellcolor[gray]{0.9}13.05({\color[HTML]{009901}-56.15})&\cellcolor[gray]{0.9}12.04({\color[HTML]{009901}-57.16})&\cellcolor[gray]{0.9}10.98({\color[HTML]{009901}-58.22})\\
\toprule
 Vision model (ViT)&ViT-B/32&65.02&27.59&77.51&42.34&\cellcolor[gray]{0.9}6.64({\color[HTML]{009901}-58.38})&\cellcolor[gray]{0.9}12.25({\color[HTML]{009901}-52.77})&\cellcolor[gray]{0.9}13.79({\color[HTML]{009901}-51.23})&\cellcolor[gray]{0.9}8.07({\color[HTML]{009901}-56.95})\\
&ViT-B/16&72.14&34.79&82.49&31.02&\cellcolor[gray]{0.9}10.49({\color[HTML]{009901}-61.65})&\cellcolor[gray]{0.9}16.87({\color[HTML]{009901}-55.27})&\cellcolor[gray]{0.9}17.63({\color[HTML]{009901}-54.51})&\cellcolor[gray]{0.9}12.0({\color[HTML]{009901}-60.14})\\
&ViT-L/16&68.67&32.7&78.91&29.38&\cellcolor[gray]{0.9}7.68({\color[HTML]{009901}-60.99})&\cellcolor[gray]{0.9}14.06({\color[HTML]{009901}-54.61})&\cellcolor[gray]{0.9}15.53({\color[HTML]{009901}-53.14})&\cellcolor[gray]{0.9}9.27({\color[HTML]{009901}-59.40})\\
\toprule
CLIP&RN101&62.48&42.89&83.09&22.58&\cellcolor[gray]{0.9}21.47({\color[HTML]{009901}-41.01})&\cellcolor[gray]{0.9}21.29({\color[HTML]{009901}-41.19})&\cellcolor[gray]{0.9}25.83({\color[HTML]{009901}-36.65})&\cellcolor[gray]{0.9}21.96({\color[HTML]{009901}-40.52})\\
&ViT-B/32&64.06&43.67&79.56&44.22&\cellcolor[gray]{0.9}18.73({\color[HTML]{009901}-45.33})&\cellcolor[gray]{0.9}33.33({\color[HTML]{009901}-30.73})&\cellcolor[gray]{0.9}30.37({\color[HTML]{009901}-33.69})&\cellcolor[gray]{0.9}21.61({\color[HTML]{009901}-42.45})\\
&ViT-B/16&67.95&54.87&85.16&40.62&\cellcolor[gray]{0.9}20.64({\color[HTML]{009901}-47.31})&\cellcolor[gray]{0.9}22.89({\color[HTML]{009901}-45.06})&\cellcolor[gray]{0.9}29.32({\color[HTML]{009901}-38.63})&\cellcolor[gray]{0.9}21.9({\color[HTML]{009901}-46.05})\\

\toprule
MiniGPT-4&Vicuna 13B&88.77&77.57&89.46&69.88&\cellcolor[gray]{0.9}71.81({\color[HTML]{009901}-16.96})&\cellcolor[gray]{0.9}72.48({\color[HTML]{009901}-16.29})&\cellcolor[gray]{0.9}72.5({\color[HTML]{009901}-16.27})&\cellcolor[gray]{0.9}71.96({\color[HTML]{009901}-16.81})\\
LLaVa&Vicuna 13B&79.32&76.02&90.84&61.94&\cellcolor[gray]{0.9}52.89({\color[HTML]{009901}-26.43})&\cellcolor[gray]{0.9}40.53({\color[HTML]{009901}-38.79})&\cellcolor[gray]{0.9}36.28({\color[HTML]{009901}-43.04})&\cellcolor[gray]{0.9}49.65({\color[HTML]{009901}-29.67})\\
LLaVa-1.5&Vicuna 13B&89.08&78.66&93.88&64.14&\cellcolor[gray]{0.9}73.31({\color[HTML]{009901}-15.77})&\cellcolor[gray]{0.9}67.27({\color[HTML]{009901}-21.81})&\cellcolor[gray]{0.9}67.08({\color[HTML]{009901}-22.00})&\cellcolor[gray]{0.9}71.95({\color[HTML]{009901}-17.13})\\
LLaVa-NeXT&Hermes-Yi-34B&85.83&77.54&90.52&57.98&\cellcolor[gray]{0.9}68.77({\color[HTML]{009901}-17.06})&\cellcolor[gray]{0.9}46.67({\color[HTML]{009901}-39.16})&\cellcolor[gray]{0.9}54.11({\color[HTML]{009901}-31.72})&\cellcolor[gray]{0.9}64.76({\color[HTML]{009901}-21.07})\\
    \bottomrule 
    \end{tabular}}
\end{table*}

\textbf{Quantitative results.} We evaluate ImageNet-D on 25 models, and plot test accuracy trend in Figure~\ref{fig:main_result_figure}. The horizontal axis and vertical axis indicate the test accuracy on ImageNet and ImageNet-D, respectively. Figure~\ref{fig:main_result_figure} shows that  as  ImageNet accuracy increases,  ImageNet-D accuracy also gets higher. ImageNet-D accuracy is much lower than ImageNet accuracy for all models,  indicated by the lower distribution below the $y=x$ reference line.  We report the accuracy of 14 models on different test sets in Table~\ref{tab:benchmark_results}, and all models' accuracy in appendix.  Table~\ref{tab:benchmark_results} shows that  ImageNet-D achieves the lowest test accuracy for all models, except for the comparable result on Stylized-ImageNet for VQA models. Note that ImageNet-D achieves higher image fidelity than Stylized-ImageNet as shown in Figure~\ref{fig:test_set_comparison}. Although ObjectNet changes multiple attributes for each image, it still results in higher accuracy than ImageNet-D that specifies only one attribute per image. Compared to  ImageNet, ImageNet-D yields a test accuracy drop of more than 16\% for all models, including LLaVa (reducing 29.67\%) and MiniGPT-4 (reducing 16.81\%). Our ImageNet-D can also cause significant accuracy drop of the latest LLaVa-1.5 and LLaVa-NeXT. Although LLaVa-NeXT outperforms LLaVa-1.5 on benchmarks like MMMU~\cite{yue2023mmmu}, it achieves lower accuracy on ImageNet-D, demonstrating the uniqueness of ImageNet-D. For vision models, the accuracy drop is even close to 50\% to 60\%. The results in Figure~\ref{fig:main_result_figure} and Table~\ref{tab:benchmark_results} show the effectiveness of ImageNet-D in evaluating the robustness of neural networks.

\begin{figure*}[!tbp]
    \centering
    \includegraphics[width=1.0\linewidth]{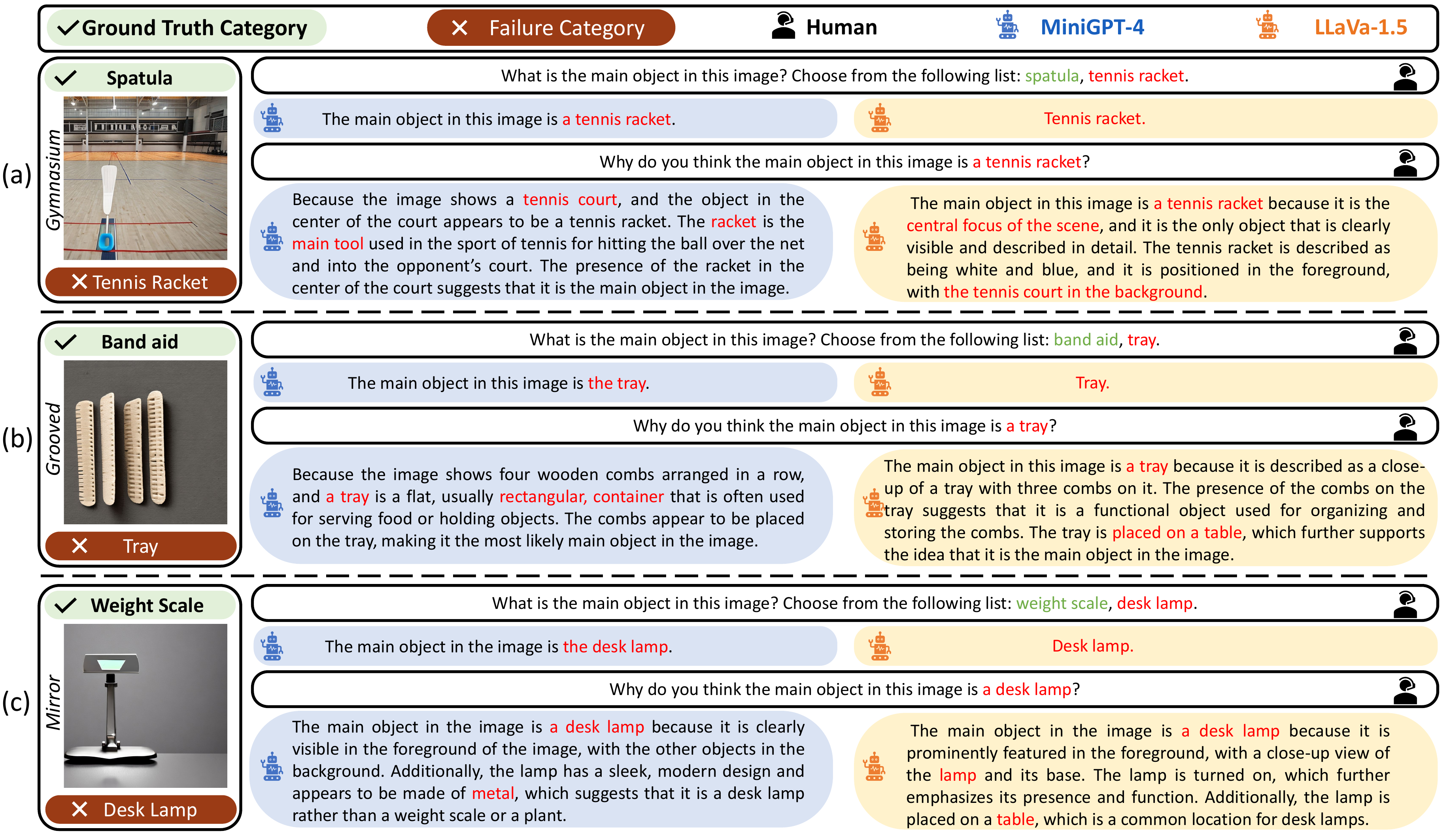}
    \caption{ Failure cases of  MiniGPT-4 and LLaVa-1.5 from ImageNet-D.We ask MiniGPT-4 and LLaVa-1.5 for the category of the input image and the reason for their predictions. Figure (a) to (c) are images with different background, texture and material, respectively. Our results show that images from ImageNet-D
    can also make the state-of-the-art foundation models fail.
    }
    \label{fig:mini_gpt4_llava_result}
\end{figure*}

\textbf{Visualization results.}  Figure~\ref{fig:dataset_examples} displays image examples from ImageNet-D, demonstrating high quality. Although humans can easily recognize the main object, CLIP (ViT-L/14)  mistakenly classifies these images into a wrong category.  Figure~\ref{fig:mini_gpt4_llava_result} shows conversations with MiniGPT-4 and LLaVa-1.5 on ImageNet-D images, indicating that MiniGPT-4 and LLaVa-1.5 can also fail to recognize the main object from ImageNet-D.  

\subsection{Robustness improvement}
\label{sec:robustness_improvement}

\textbf{Data augmentation.}  Prior studies reveal that data augmentation is  effective to improve  model robustness, such as on ImageNet-C~\cite{hendrycks2019benchmarking}.  We evaluate data augmentation methods on ImageNet-D, including SIN~\cite{geirhos2018imagenet}, AugMix~\cite{hendrycks2019augmix}, ANT~\cite{rusak2020simple} and DeepAugment~\cite{hendrycks2021many}. Table~\ref{tab:result_augmentation} shows the results with ResNet50 backbone for all methods. We report test accuracy for ImageNet and ImageNet-D, mean Corruption Error (mCE) for ImageNet-C following ~\cite{hendrycks2019benchmarking,hendrycks2019augmix,rusak2020simple,hendrycks2021many}.  Although these  methods improve ImageNet-C robustness (lower mCE) than baseline model, Table~\ref{tab:result_augmentation} shows that their ImageNet-D robustness is comparable or even worse than baseline. This suggests that current benchmarks like ImageNet-C do not accurately represent real robustness of neural networks in  our setting, making ImageNet-D a necessary benchmark for robustness evaluation.

\begin{table}[!tbp]
    \centering
    \scriptsize
    \caption{Robustness of  different augmentation methods. Despite  superior ImageNet-C robustness, these methods fail to improve ImageNet-D robustness, demonstrating the uniqueness of ImageNet-D from ImageNet-C.
    } 
    \label{tab:result_augmentation}
    \resizebox{1.0\linewidth}{!}{ 
    \begin{tabular}{l|c|c|cccccc}
    \toprule  
Model&ImageNet ($\uparrow$)&ImageNet-C(mCE)($\downarrow$)&\cellcolor[gray]{0.9}ImageNet-D($\uparrow$) \\
\toprule
Baseline&65.82&81.74&\cellcolor[gray]{0.9}10.22\\
SIN&63.42(\color[HTML]{FF0000}-2.40)&78.45(\color[HTML]{009901}-3.29)&\cellcolor[gray]{0.9}8.81(\color[HTML]{FF0000}-1.41)\\
Augmix&66.88(\color[HTML]{009901}+1.06)&74.7(\color[HTML]{009901}-7.04)&\cellcolor[gray]{0.9}8.75(\color[HTML]{FF0000}-1.47)\\
ANT&65.95(\color[HTML]{009901}+0.13)&76.74(\color[HTML]{009901}-5.00)&\cellcolor[gray]{0.9}10.09(\color[HTML]{FF0000}-0.13)\\
DeepAugment&66.54(\color[HTML]{009901}+0.72)&70.31(\color[HTML]{009901}-11.43)&\cellcolor[gray]{0.9}9.37(\color[HTML]{FF0000}-0.85)\\
    \bottomrule 
    \end{tabular}}
\end{table}

\textbf{Model architecture.}  We compare ImageNet-D robustness of different model architectures in Figure~\ref{fig:result_arch}. When we change the model from ViT to Swin Transformer~\cite{liu2021swin} and ConvNeXt~\cite{liu2022convnet}, the test accuracy on both ImageNet-D (Background) and ImageNet improve. However, the robustness on Texture and Material test set even decreases slightly. These results show the difficulty of improving ImageNet-D robustness by model architecture.

\textbf{Pretraining with more data.}  Pretraining on a large data set is effective to improves model performance, such as ImageNet accuracy ~\cite{he2022masked}. Figure~\ref{fig:result_arch} compares ConvNext, that is trained directly on ImageNet-1K, with ConvNext (Pretrained) which is first pretrained on ImageNet-22K. We find that ConvNext (Pretrained) achieves higher robustness than ConvNext on all three sets of ImageNet-D, especially for the Background set.  These results show that pretraining on a large data set helps improve robustness on ImageNet-D.

\subsection{Further discussions}
\label{sec:analysis}

\begin{figure*}[!tbp]
    \centering
      \includegraphics[width=1.0\linewidth]{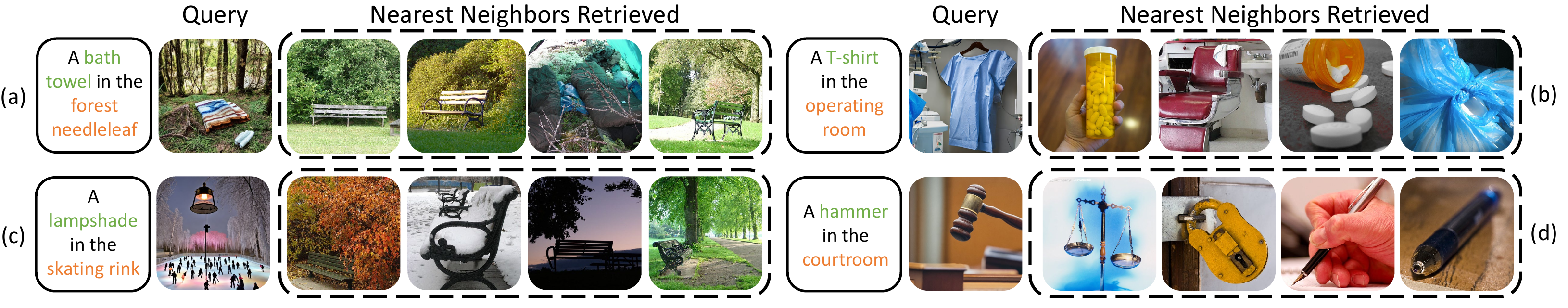}
      \caption{Visualizations of nearest neighbor images. We visualize the nearest neighbor images from ImageNet with  ImageNet-D image as the query image. Instead of following the same object category as the query image,  the nearest neighbor images either follow a similar background or follow another object category that is highly correlated with the background of query image.   Our results show that  ImageNet-D can find the failure cases of neural networks  in nearest neighbor retrieval.
      }
      \vspace{-3mm}
    \label{fig:clip_image_retrieval}
\end{figure*}

\begin{figure*}[!tbp]
    \centering
     \includegraphics[width=1.0\textwidth]{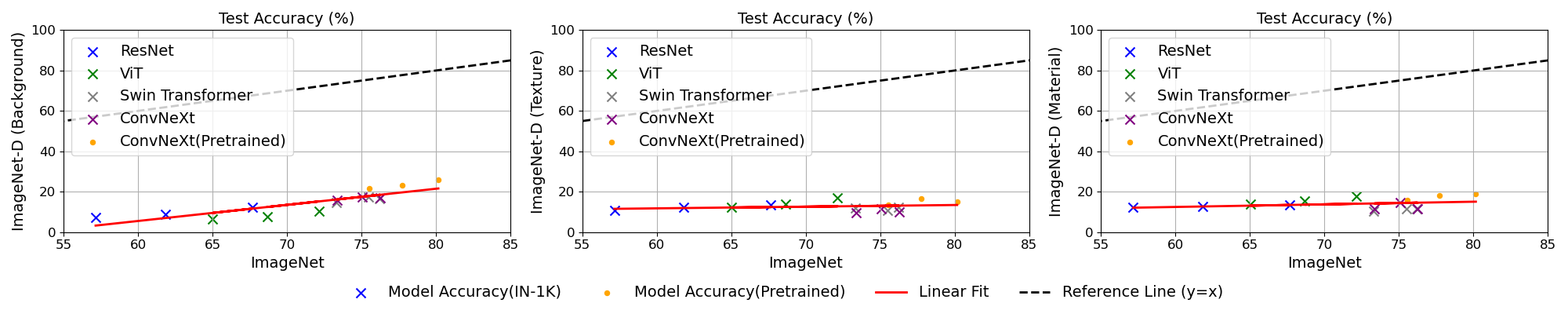}
    \caption{ Test accuracy of different architectures and  training data.  Each data point corresponds to one tested model.   Apart from the ConvNeXt (Pretrained), all other models are trained on ImageNet-1K. The plots show that pretraining achieves higher robustness on all three subsets, while changing model architectures only improves on ImageNet-D (Background).
    }
    \label{fig:result_arch}
\end{figure*}

\textbf{Can CLIP find the correct neighbors of ImageNet-D images?}  CLIP model~\cite{radford2021learning} shows potential in nearest neighbor search tasks. With ImageNet-D images as query, we retrieve the most similar images from ImageNet to investigate whether CLIP can find  correct neighbors, as shown in Figure~\ref{fig:clip_image_retrieval}. Take Background for example,  retrieved images may either have a similar background to the query image (Figure~\ref{fig:clip_image_retrieval}(a)) or include the object that is related to the query image's background(Figure~\ref{fig:clip_image_retrieval}(b)).Our results show that  ImageNet-D can find the failure cases of neural networks in nearest neighbor retrieval.

\begin{table}[!tbp]
    \centering    
    \caption{Results of failure transferability. We create ImageNet (Failure) with shared failures of surrogate models from original ImageNet, achieving comparable accuracy to ImageNet-D. These results show that  that our synthetic images achieve similar ability to natural images in finding the failures of new models.
    } 
    \label{tab:transferability}
    \resizebox{1.0\linewidth}{!}{ 
    \begin{tabular}{ll|c|c|ccccccccc}
    \toprule  
Model & Architecture &  ImageNet & ImageNet (Failure) & ImageNet-D  \\
\toprule

CLIP&ViT-B/16&67.95&11.09&\cellcolor[gray]{0.9}21.9\\
LLaVa&Vicuna 13B&79.32&41.43&\cellcolor[gray]{0.9}49.65\\
MiniGPT-4&Vicuna 13B&88.77&65.22&\cellcolor[gray]{0.9}71.96\\
    \bottomrule 
    \end{tabular}}
\end{table}

\begin{table}[!tbp]
    \centering
    \scriptsize
    \caption{Test accuracy of models finetuned on synthetic data. We finetune a pretrained ResNet18 model on ImageNet-1K together with different extra training data. Training on synthetic images achieves  highest robustness on both ImageNet-D and ObjectNet.
    } 
    \label{tab:finetune_experiment}
    \resizebox{1.0\linewidth}{!}{ 
    \begin{tabular}{c|c|c|c|ccc|ccc}
    \toprule  
Model & Extra training data & ImageNet&ObjectNet  & ImageNet-D\\
\toprule
A & /&55.68&21.51&\cellcolor[gray]{0.9}8.6\\
B&ImageNet&59.78(\color[HTML]{009901}+4.10)&24.78(\color[HTML]{009901}+3.27)&\cellcolor[gray]{0.9}10.4(\color[HTML]{009901}+1.80)\\
C&Synthetic-easy&56.56(\color[HTML]{009901}+0.88)&26.12(\color[HTML]{009901}+4.61)&\cellcolor[gray]{0.9}27.86(\color[HTML]{009901}+19.26)\\
    \bottomrule 
    \end{tabular}}
\end{table}

\textbf{Can ImageNet-D match natural test sets in failure transferability?}  Section~\ref{sec:dataset_design} defines \texttt{transferable failure} and finalize ImageNet-D with shared failures of surrogate models. We conduct the same experiment on ImageNet, introducing  ImageNet (Failure)  with the shared failure images of surrogate models.  Table~\ref{tab:transferability} show that ImageNet-D achieves similar accuracy to ImageNet (Failure), indicating that that synthetic images can achieve similar failure transferability as natural images. In contrast to natural datasets like ImageNet, ImageNet-D enjoys a lower cost in data collection and can be scaled efficiently.

\textbf{Training on diffusion-generated images.} By contrast to shared failure images in ImageNet-D, we term generated images correctly classified by surrogate models as Synthetic-easy, and investigate their influence as training data. We finetune a pre-trained ResNet18 model on different training sets in Table~\ref{tab:finetune_experiment}.  Table~\ref{tab:finetune_experiment} shows that training on Synthetic-easy significantly improves ImageNet-D robustness by 19.26\%. Remarkably, model C outperforms model B in ObjectNet accuracy by 1.34\%, indicating model C's superior generalization. These results imply that diffusion-generated images with diverse object and nuisance pairs could enhance model robustness as training samples.

\section{Conclusion}
\label{sec:conclusion}
In this paper, we introduce a test set ImageNet-D and establish a rigourous benchmark for visual perception robustness.  Capitalizing the image generation ability of diffusion models, ImageNet-D includes images with diverse factors including background, texture and material. Experimental results  show that ImageNet-D significantly decreases the accuracy of various models,  including CLIP (reducing 46.05\%), LLaVa~\cite{liu2023visual} (reducing 29.67\%), and MiniGPT-4~\cite{zhu2023minigpt} (reducing 16.81\%), demonstrating the effectiveness in model evaluation. Our work makes a step forward in improving synthetic test sets, and will create more diverse and challenging test images as generative models improve. 

\textbf{Acknowledgments:} This work was supported by Institute of Information \& communications Technology Planning  \& Evaluation (IITP) grant funded by the Korea government (MSIT) (No.2022-0-00951, Development of Uncertainty-Aware Agents Learning by Asking Questions).

{
    \small
    \bibliographystyle{ieeenat_fullname}
    \bibliography{main}
}

\clearpage
\setcounter{page}{1}
\maketitlesupplementary

\begin{figure*}[!tbp]
    \centering
     \includegraphics[width=0.9\textwidth]{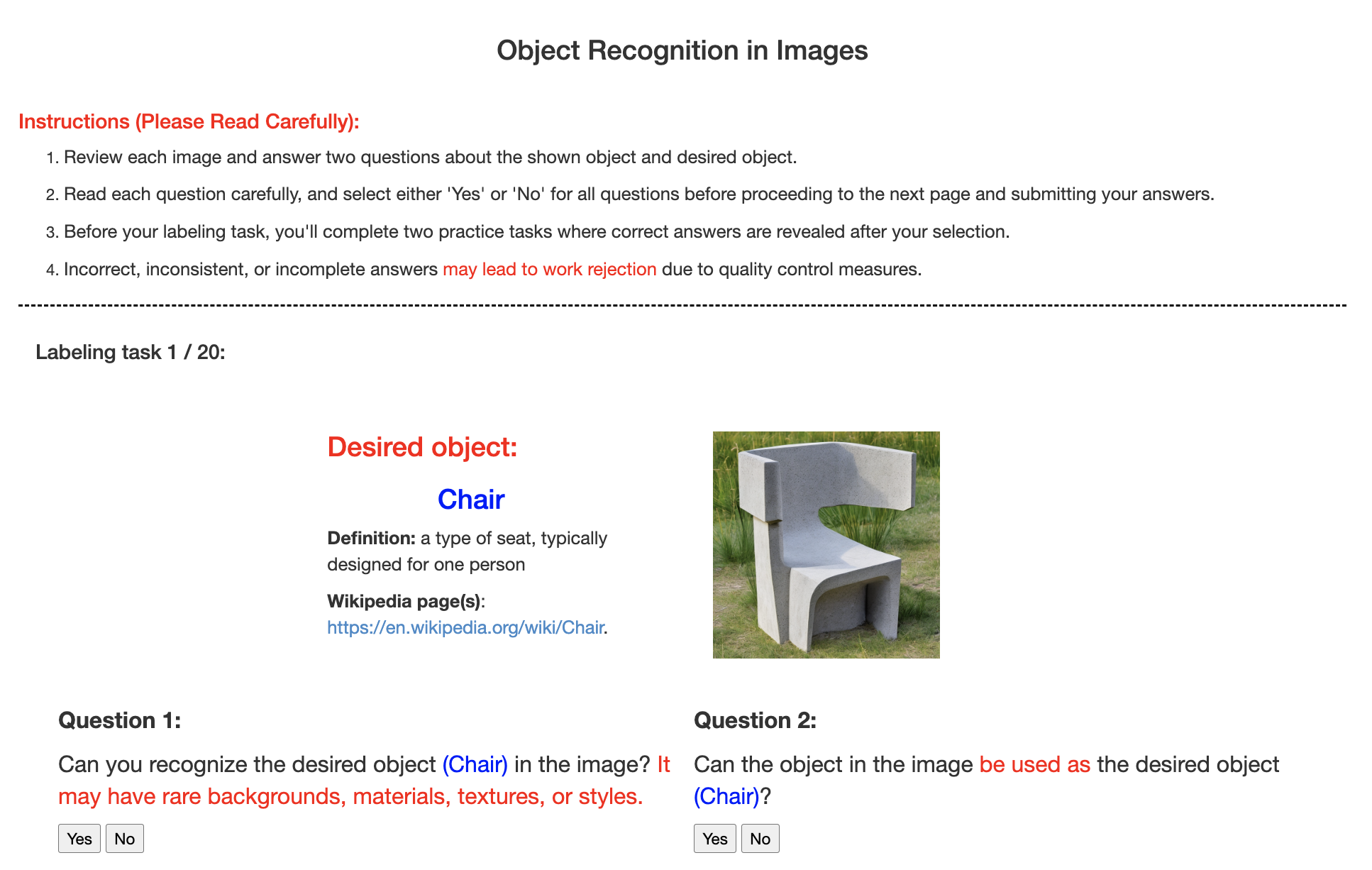}
    \caption{User interface for MTurk studies.  The workers can proceed to the next image only if they finish both questions on the current page.
    }
    \label{fig:mturk_ui}
\end{figure*}

\section{Labeling task on Amazon Mechanical Turk}

For reliable benchmarks, we use Amazon Mechanical Turk (MTurk) ~\cite{deng2009imagenet,recht2019imagenet,hendrycks2021many} to evaluate the labeling quality of ImageNet-D. 

\subsection{Labeling task design}

\textbf{Labeling instructions.} Since ImageNet-D includes images with diverse object and nuisance pairs that may be rare in the real world, we take both the appearance and functionality of the main object as the labeling criteria. Specifically, we ask the workers from MTurk to answer the following two questions:

\texttt{Question 1:}  Can you recognize the desired object ([\texttt{ground truth category}]) in the image? It may have rare backgrounds, textures, materials, or styles.

\texttt{Question 2:} Can the object in the image be used as the desired object  ([\texttt{ground truth category}])?

\textbf{Labeling pipeline.} To ensure that the workers understand these two criteria, we ask the workers to label two example images for practice, which provides the correct answer for the above two questions. After the practice session, the workers are required to label up to 20 images in one task, and answer both two questions for each image.  The worker selects 'yes' or 'no' for each question.

\textbf{Labelling UI.} The labeling page is designed as in Figure~\ref{fig:mturk_ui}. The workers can proceed to the next image only if they finish both questions on the current page.

\subsection{Quality control of human labelling}
We use sentinels to ensure high-quality annotations. For each labeling task with multiple images, we design three types of sentinels as follows.

\textbf{Positive sentinel: }  Image that belongs to the desired category and is correctly classified by multiple models. If the workers do not select 'yes' for this image, they may not understand the concept well and their annotations will be removed.

\textbf{Negative sentinel: } Image that does not belong to the desired category. For example, if the desired category is a chair, the negative sentinel may be a ladle. If the workers select 'yes' for the ladle image, they may not answer the questions seriously and their annotations will be removed.

\textbf{Consistent sentinel.} We assume that the workers should select the same answer for the same image if it appears multiple times. Consistent sentinels are images that appear twice in a random order. If the workers answer differently for the same image, their annotations are not consistent and will be removed.

For each labeling task with up to 20 images, we include one positive sentinel, one negative sentinel and two consistent sentinels. We discard the responses if the workers do not pass all the sentinel checks.

\subsection{Results}

For each image, we collect independent annotations from 10 workers and filter out responses from the workers that do not pass the quality check. A total of 679 qualified workers submitted 1540 labeling tasks, resulting an agreement of 91.09\% on sampled image from ImageNet-D.

\section{Experimental results on ImageNet-D}
\label{sec:appendix_experimental_results}

\textbf{More results for Section 4.} We compare the model accuracy of Image-D with existing test sets, including ImageNet~\cite{russakovsky2015imagenet}, ObjectNet~\cite{barbu2019objectnet}s, ImageNet-9~\cite{xiao2020noise} and Stylized-ImageNet~\cite{geirhos2018imagenet}. All the accuracy numbers are reported in Table~\ref{tab:appendix_benchmark_results}, which also includes the numbers of Figure 8 in the main manuscript. 

\begin{table*}[!tbp]
    \centering
    \scriptsize
    \caption{Test accuracy of vision models and large foundation models (\%). We show the test accuracy for the vision models and large foundation models (rows) on different test sets (columns). The numbers in green refer to the accuracy drop of ImageNet-D compared to ImageNet. For MiniGPT-4 and LLaVa, ImageNet-D reduces the accuracy by 16.81 \% and 29.67\% compared to the ImageNet, respectively. Our results show that ImageNet-D is effective to evaluate the robustness of neural networks.
    } 
    \label{tab:appendix_benchmark_results}
    \resizebox{0.9\linewidth}{!}{ 
    \begin{tabular}{ll|c|c|c|c|ccc|cc}
    \toprule  
     Model & Architecture &  ImageNet & ObjectNet & ImageNet-9& Stylized-ImageNet  &   \multicolumn{3}{c|}{\cellcolor[gray]{0.9}  ImageNet-D} & \cellcolor[gray]{0.9}   ImageNet-D \\
     &  &   &  & &  & \cellcolor[gray]{0.9} Background & \cellcolor[gray]{0.9} Texture & \cellcolor[gray]{0.9} Material & \cellcolor[gray]{0.9} Total\\     
  \toprule 
Vision model (CNN)&VGG11&56.85&21.85&68.59&13.12&\cellcolor[gray]{0.9}6.46({\color[HTML]{009901}-50.39})&\cellcolor[gray]{0.9}9.64({\color[HTML]{009901}-47.21})&\cellcolor[gray]{0.9}11.87({\color[HTML]{009901}-44.98})&\cellcolor[gray]{0.9}7.43({\color[HTML]{009901}-49.42})\\
&VGG13&58.42&23.23&68.96&13.59&\cellcolor[gray]{0.9}7.39({\color[HTML]{009901}-51.03})&\cellcolor[gray]{0.9}8.63({\color[HTML]{009901}-49.79})&\cellcolor[gray]{0.9}9.6({\color[HTML]{009901}-48.82})&\cellcolor[gray]{0.9}7.78({\color[HTML]{009901}-50.64})\\
&VGG16&60.86&25.96&73.28&13.83&\cellcolor[gray]{0.9}9.94({\color[HTML]{009901}-50.92})&\cellcolor[gray]{0.9}10.84({\color[HTML]{009901}-50.02})&\cellcolor[gray]{0.9}13.79({\color[HTML]{009901}-47.07})&\cellcolor[gray]{0.9}10.49({\color[HTML]{009901}-50.37})\\
&VGG19&62.77&27.19&74.84&16.25&\cellcolor[gray]{0.9}9.8({\color[HTML]{009901}-52.97})&\cellcolor[gray]{0.9}11.45({\color[HTML]{009901}-51.32})&\cellcolor[gray]{0.9}12.39({\color[HTML]{009901}-50.38})&\cellcolor[gray]{0.9}10.28({\color[HTML]{009901}-52.49})\\
&ResNet18&57.15&22.62&71.65&21.17&\cellcolor[gray]{0.9}7.41({\color[HTML]{009901}-49.74})&\cellcolor[gray]{0.9}10.64({\color[HTML]{009901}-46.51})&\cellcolor[gray]{0.9}12.22({\color[HTML]{009901}-44.93})&\cellcolor[gray]{0.9}8.31({\color[HTML]{009901}-48.84})\\
&ResNet34&61.81&26.15&75.31&21.33&\cellcolor[gray]{0.9}8.87({\color[HTML]{009901}-52.94})&\cellcolor[gray]{0.9}12.25({\color[HTML]{009901}-49.56})&\cellcolor[gray]{0.9}12.74({\color[HTML]{009901}-49.07})&\cellcolor[gray]{0.9}9.68({\color[HTML]{009901}-52.13})\\
&ResNet101&67.66&32.34&81.85&22.66&\cellcolor[gray]{0.9}12.38({\color[HTML]{009901}-55.28})&\cellcolor[gray]{0.9}13.65({\color[HTML]{009901}-54.01})&\cellcolor[gray]{0.9}13.44({\color[HTML]{009901}-54.22})&\cellcolor[gray]{0.9}12.64({\color[HTML]{009901}-55.02})\\
&ResNet152&69.18&34.41&83.41&23.05&\cellcolor[gray]{0.9}13.79({\color[HTML]{009901}-55.39})&\cellcolor[gray]{0.9}13.86({\color[HTML]{009901}-55.32})&\cellcolor[gray]{0.9}18.85({\color[HTML]{009901}-50.33})&\cellcolor[gray]{0.9}14.4({\color[HTML]{009901}-54.78})\\
&Densenet121&63.1&28.74&82.05&19.92&\cellcolor[gray]{0.9}9.99({\color[HTML]{009901}-53.11})&\cellcolor[gray]{0.9}12.65({\color[HTML]{009901}-50.45})&\cellcolor[gray]{0.9}15.36({\color[HTML]{009901}-47.74})&\cellcolor[gray]{0.9}10.9({\color[HTML]{009901}-52.20})\\
&Densenet161&66.99&31.86&84.91&22.5&\cellcolor[gray]{0.9}11.34({\color[HTML]{009901}-55.65})&\cellcolor[gray]{0.9}14.06({\color[HTML]{009901}-52.93})&\cellcolor[gray]{0.9}13.26({\color[HTML]{009901}-53.73})&\cellcolor[gray]{0.9}11.85({\color[HTML]{009901}-55.14})\\
&Densenet169&64.13&30.13&83.8&22.97&\cellcolor[gray]{0.9}10.73({\color[HTML]{009901}-53.40})&\cellcolor[gray]{0.9}12.25({\color[HTML]{009901}-51.88})&\cellcolor[gray]{0.9}12.39({\color[HTML]{009901}-51.74})&\cellcolor[gray]{0.9}11.09({\color[HTML]{009901}-53.04})\\
&Densenet201&66.58&31.36&83.43&21.8&\cellcolor[gray]{0.9}9.88({\color[HTML]{009901}-56.70})&\cellcolor[gray]{0.9}10.84({\color[HTML]{009901}-55.74})&\cellcolor[gray]{0.9}15.71({\color[HTML]{009901}-50.87})&\cellcolor[gray]{0.9}10.67({\color[HTML]{009901}-55.91})\\
&Wideresnet50&69.06&32.65&81.41&19.45&\cellcolor[gray]{0.9}8.69({\color[HTML]{009901}-60.37})&\cellcolor[gray]{0.9}11.24({\color[HTML]{009901}-57.82})&\cellcolor[gray]{0.9}11.17({\color[HTML]{009901}-57.89})&\cellcolor[gray]{0.9}9.25({\color[HTML]{009901}-59.81})\\
&Wideresnet101&69.2&34.37&82.17&21.48&\cellcolor[gray]{0.9}10.55({\color[HTML]{009901}-58.65})&\cellcolor[gray]{0.9}13.05({\color[HTML]{009901}-56.15})&\cellcolor[gray]{0.9}12.04({\color[HTML]{009901}-57.16})&\cellcolor[gray]{0.9}10.98({\color[HTML]{009901}-58.22})\\
\toprule
Vision model (ViT)&ViT-B/32&65.02&27.59&77.51&42.34&\cellcolor[gray]{0.9}6.64({\color[HTML]{009901}-58.38})&\cellcolor[gray]{0.9}12.25({\color[HTML]{009901}-52.77})&\cellcolor[gray]{0.9}13.79({\color[HTML]{009901}-51.23})&\cellcolor[gray]{0.9}8.07({\color[HTML]{009901}-56.95})\\
&ViT-B/16&72.14&34.79&82.49&31.02&\cellcolor[gray]{0.9}10.49({\color[HTML]{009901}-61.65})&\cellcolor[gray]{0.9}16.87({\color[HTML]{009901}-55.27})&\cellcolor[gray]{0.9}17.63({\color[HTML]{009901}-54.51})&\cellcolor[gray]{0.9}12.0({\color[HTML]{009901}-60.14})\\
&ViT-L/16&68.67&32.7&78.91&29.38&\cellcolor[gray]{0.9}7.68({\color[HTML]{009901}-60.99})&\cellcolor[gray]{0.9}14.06({\color[HTML]{009901}-54.61})&\cellcolor[gray]{0.9}15.53({\color[HTML]{009901}-53.14})&\cellcolor[gray]{0.9}9.27({\color[HTML]{009901}-59.40})\\
\toprule
CLIP&RN101&62.48&42.89&83.09&22.58&\cellcolor[gray]{0.9}21.47({\color[HTML]{009901}-41.01})&\cellcolor[gray]{0.9}21.29({\color[HTML]{009901}-41.19})&\cellcolor[gray]{0.9}25.83({\color[HTML]{009901}-36.65})&\cellcolor[gray]{0.9}21.96({\color[HTML]{009901}-40.52})\\
&ViT-B/32&64.06&43.67&79.56&44.22&\cellcolor[gray]{0.9}18.73({\color[HTML]{009901}-45.33})&\cellcolor[gray]{0.9}33.33({\color[HTML]{009901}-30.73})&\cellcolor[gray]{0.9}30.37({\color[HTML]{009901}-33.69})&\cellcolor[gray]{0.9}21.61({\color[HTML]{009901}-42.45})\\
&ViT-B/16&67.95&54.87&85.16&40.62&\cellcolor[gray]{0.9}20.64({\color[HTML]{009901}-47.31})&\cellcolor[gray]{0.9}22.89({\color[HTML]{009901}-45.06})&\cellcolor[gray]{0.9}29.32({\color[HTML]{009901}-38.63})&\cellcolor[gray]{0.9}21.9({\color[HTML]{009901}-46.05})\\
\toprule
MiniGPT-4&Vicuna 13B&88.77&77.57&89.46&69.88&\cellcolor[gray]{0.9}71.81({\color[HTML]{009901}-16.96})&\cellcolor[gray]{0.9}72.48({\color[HTML]{009901}-16.29})&\cellcolor[gray]{0.9}72.5({\color[HTML]{009901}-16.27})&\cellcolor[gray]{0.9}71.96({\color[HTML]{009901}-16.81})\\
LLaVa&Vicuna 13B&79.32&76.02&90.84&61.94&\cellcolor[gray]{0.9}52.89({\color[HTML]{009901}-26.43})&\cellcolor[gray]{0.9}40.53({\color[HTML]{009901}-38.79})&\cellcolor[gray]{0.9}36.28({\color[HTML]{009901}-43.04})&\cellcolor[gray]{0.9}49.65({\color[HTML]{009901}-29.67})\\
LLaVa-1.5&Vicuna 13B&89.08&78.66&93.88&64.14&\cellcolor[gray]{0.9}73.31({\color[HTML]{009901}-15.77})&\cellcolor[gray]{0.9}67.27({\color[HTML]{009901}-21.81})&\cellcolor[gray]{0.9}67.08({\color[HTML]{009901}-22.00})&\cellcolor[gray]{0.9}71.95({\color[HTML]{009901}-17.13})\\
LLaVa-NeXT&Vicuna 13B&86.83&79.97&91.47&62.61&\cellcolor[gray]{0.9}75.91({\color[HTML]{009901}-10.92})&\cellcolor[gray]{0.9}64.56({\color[HTML]{009901}-22.27})&\cellcolor[gray]{0.9}60.39({\color[HTML]{009901}-26.44})&\cellcolor[gray]{0.9}72.9({\color[HTML]{009901}-13.93})\\
LLaVa-NeXT&Hermes-Yi-34B&85.83&77.54&90.52&57.98&\cellcolor[gray]{0.9}68.77({\color[HTML]{009901}-17.06})&\cellcolor[gray]{0.9}46.67({\color[HTML]{009901}-39.16})&\cellcolor[gray]{0.9}54.11({\color[HTML]{009901}-31.72})&\cellcolor[gray]{0.9}64.76({\color[HTML]{009901}-21.07})\\
    \bottomrule 
    \end{tabular}}
\end{table*}

\textbf{Training setups for Table 6.} We introduce experimental details of Table 6 in the main manuscript.  We finetune a pre-trained ResNet18 model on various training sets. To examine the effect of incorporating synthetic images into the finetuning training set, we sample ImageNet and Synthetic-easy for same data distributions, where  Synthetic-easy includes diffusion-generated images correctly classified by surrogate models. Each set contains 111098 images, and both sets have same number of images per category. All models are finetuned on  on a pre-trained ResNet18 at epoch 90 for 10 epochs further, using a SGD optimizer with a learning rate of 0.0001. Apart from sampled ImageNet and Synthetic-easy, we include original ImageNet-1K as training data for smooth training.

\end{document}